\documentclass[a4paper,fleqn]{cas-dc}

\usepackage[numbers]{natbib}
\usepackage{graphics} 
\usepackage{graphicx}
\usepackage{array}
\usepackage{ragged2e}
\usepackage{float}
\usepackage{wrapfig}
\usepackage{subfigure}
\usepackage{textcomp}
\usepackage{verbatim}

\usepackage{url}
\usepackage{hyperref}
\usepackage{amsmath}
\usepackage{algpseudocode}
\usepackage{amssymb}
\usepackage{enumitem}
\usepackage{algorithm}
\usepackage{hhline}
\usepackage{hyperref}
\hypersetup{%
  colorlinks=true,
  citecolor={blue}
}

\begin{document}
\let\WriteBookmarks\relax
\def\floatpagepagefraction{1}
\def\textpagefraction{.001}
\shorttitle{A Deep Learning Approach towards Quanitfying Marine Plastic}
\shortauthors{Tata et~al.}

\title [mode = title]{A Robotic Approach towards Quantifying Epipelagic Bound Plastic Using Deep Visual Models
}                      

\tnotetext[1]{This document is the results of the research
   project funded by The Ocean Cleanup.}

\author[1]{Gautam Tata}
\cormark[1]
\fnmark[1, 3]
\ead{gtata@csumb.edu}

\credit{Conceptualization of this study, Methodology, Software, Validation, Investigation, Resources, Data Curation, Writing - Original Draft, Supervision}

\address[1]{California State University, Monterey Bay}

\author[1]{Jay Lowe}
\fnmark[2]
\ead{jlowe@csumb.edu}

\credit{Software, Validation, Data Curation, Writing - Review and Editing, Project administration}

\author[2]{Olivier Poirion}
\fnmark[2]
\ead{opoirion@ucsd.edu }
\address[2]{The Jackson Laboratory, Connecticut, United States of America
}

\credit{Writing - Review and Editing, Resources}

\author%
[3]
{Sarah-Jeanne Royer}
\cormark[3]
\fnmark[1,3]
\ead{sroyer@hpu.edu}

\address[3]{Scripps Institution of Oceanography, San Diego, California, United States of America 
}
\address[3]{The Ocean Cleanup Foundation, Rotterdam, The Netherlands}
\address[3]{Center for Marine Debris Research, Hawaii Pacific University, Waimanalo 96795, Hawaii, United States of America}

\credit{Supervision, Project administration, Writing - Review and Editing, Funding acquisition, Resources}

\cortext[cor1]{Principal Author}
\cortext[cor2]{corresponding author}

\begin{abstract}
The quantification of positively buoyant marine plastic debris is critical to understanding how plastic litter accumulates across the world's oceans and is also crucial to identifying hotspots for targeted cleanup efforts. Currently, the most common method to quantify marine plastic is using manta trawls for manual sampling. However, this method is cost-intensive and requires human labor. This study removes the need for manual sampling by using an autonomous method using neural networks and computer vision models, which trained on images captured from various layers of the ocean column to perform real-time plastic quantification. The best performing model has a Mean Average Precision of 85\% and an F1-Score of 0.89 while maintaining near real-time processing speeds ~2 ms/img.

\end{abstract}





\maketitle

\section{Introduction}

Plastic pollution poses an imminent threat to the marine environment, food safety \cite{BARBOZA2018336}, human health, eco-tourism, and contributes to climate change \cite{schmidt2017export}. Global plastic production has exceeded 500 million tons of plastic, and projections indicate that 30\% of all produced plastic will end up discarded in the oceans \cite{nollkaemper1994land} \cite{epa2014municipal}. Researchers have documented a five-fold increase in plastic debris within the Central Pacific Gyre and have shown that plastic pieces now outnumber the native plankton 6:1 in terms of abundance \cite{clapp2012rising}.
A significant amount of marine plastic (about 80\%) originates from land-based sources \cite{Windom_1992}: Most commonly in the form of food containers, such as plastic bags and bottles, and packaging materials. The other ~20\% stems from shipping vessel discharges and discarded commercial fishing gear \cite{Windom_1992}. 
Studies have shown that removing plastic from the oceans will exponentially benefit the ecosystems. This includes the prevention of the movement of invasive species between regions \cite{carlton2017tsunami}, the prevention of its degradation into micro-plastics \cite{andrady2011microplastics}, and the decrease in emissions of greenhouse gases (thereby decelerating climate change) \cite{royer2018production}]. 
To understand the spatiotemporal distribution of plastic, we require more accurate methods with reliable and low-cost deployment strategies. Various in situ approaches to ocean plastic monitoring have been proposed. These in situ methods include using SONAR/LIDAR to map plastic debris \cite{valdenegrotoro2019deep}, human counting via visual methods \cite{van2018methodology}, and debris sampling using fishing nets \cite{rech2014rivers}. However, these methods are labor-intensive, incur high financial costs, and do not cover large surface areas. 
Furthermore, polymers such as polyethylene and polypropylene are affected by the growth of a biofilm when submerged in water that will influence their sinking behaviors \cite{Kaiser_Kowalski_Waniek_2017}. Any polymer that has its density increased by biofilm to a certain point sinks beyond surface sampling devices such as manta trawls. As a result, these surface sampling limitations lead to quantity underestimations of floating plastics.
Creating an accurate marine plastic debris estimation requires developing alternative methods to investigate the distribution of positively buoyant plastic across the entire water column. 
Recently, several methods using computer vision and modern deep learning technologies to quantify marine plastic debris without physical removal have been suggested \cite{oceancleanup} \cite{fulton2018robotic}.
The Earth and Space Science journal illustrates a method using the two-stage Faster R-CNN model to actively monitor and identify surface plastic as it floats down a river \cite{oceancleanup}. This approach does not account for the sinking polymer problem but shows that, on average, an automated method detects 34.6\% more plastic than human visual counting does.
Remote sensing of plastic litter provides a promising new and less labor-intensive tool for the quantification and characterization of ocean plastic pollution \cite{rs13173401}. A research team at the University of Minnesota developed a computer vision model specialized for marine plastic detection in deep-sea environments \cite{fulton2018robotic} which demonstrates that quantification across the water column can be achieved. It also exemplifies the relationship between object detection models and AUV’s to great success.
The AquaVision project \cite{PANWAR2020100026} shows that object detection models can reach high levels of precision utilizing open-source datasets and one-stage approaches such as RetinaNet. Since AquaVision was trained on the TACO dataset. It also indicates that a computer vision model trained on land-based images of plastic can detect similar types of plastic in a marine environment.
Unlike these recently proposed algorithms that specialize in monitoring either floating marine plastic \cite{oceancleanup}, deep-sea specific environments \cite{fulton2018robotic}, or models trained on land-based plastic [17]. The Ocean Cleanup group has demonstrated that computer vision models can detect floating plastic debris via cameras attached to above-water vessels \cite{rs13173401}. Their results show that macroplastics can be successfully quantified for comparisons across methods. Our object detection model (DeepPlastic) utilizes a training set composed exclusively of marine-based plastic images and performs equally well across the entire water column while producing significant results.

\begin{figure}[ht]
\centering
\includegraphics[width=0.90\linewidth]{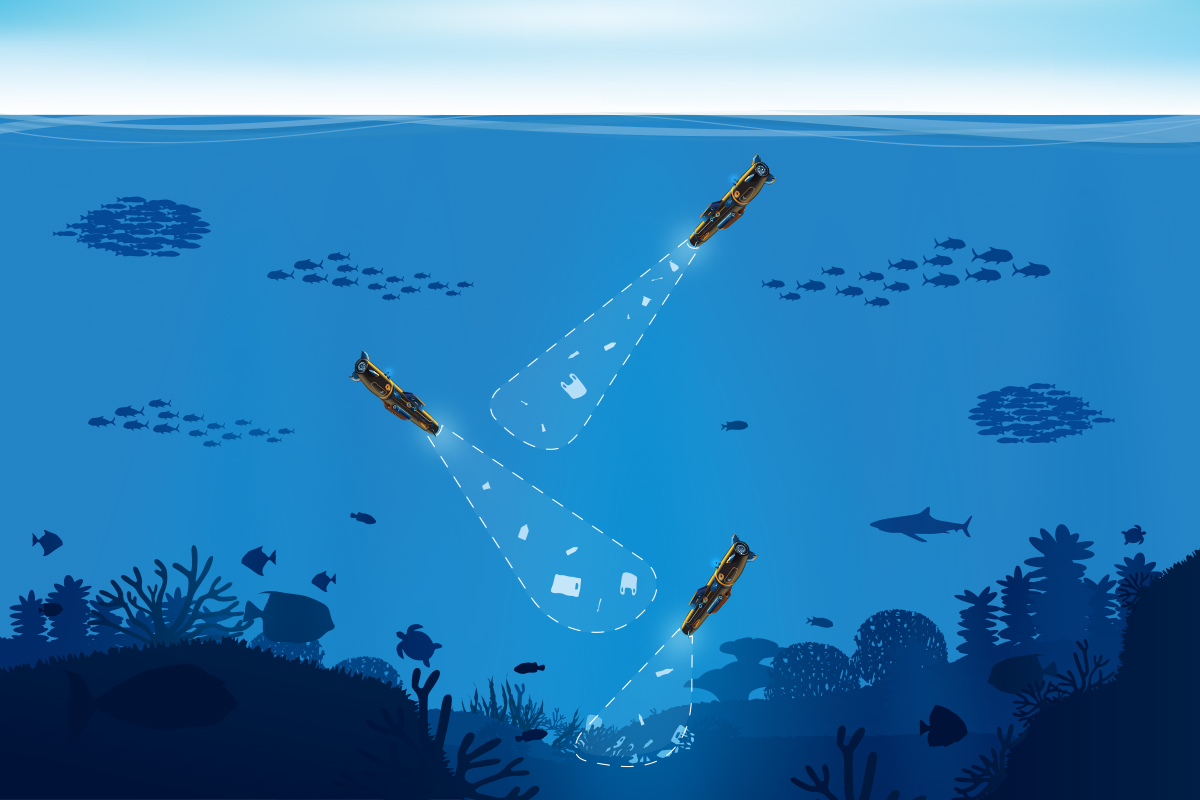}
\caption{Concept of real-time plastic detection via AUV's equipped with cameras and DeepTrash vision}
\label{fig_concept}
\end{figure}

In this study, we tested four state-of-the-art deep-learning architectures, Faster-RCNN, Single Shot Multibox Detector, YOLOv4-Tiny and YOLOv5-S, then reported their performances to infer marine plastic debris in real-time. The main results will be described as follows: 1) the model’s precision and accuracy to feasibly identify plastic debris, 2) insurability that this method can successfully distinguish marine plastic debris from similar-looking non-plastic objects, and 3) A generalized model capable of detecting marine plastic in most oceanic environments. The results show that deep learning models can identify plastic with significant accuracy while operating at a rate that supports real-time applications such as autonomous underwater vehicles (AUVs) for at-scale marine- plastic quantification and monitoring.

\section{Related Work}
Increasing demand for identifying and removing plastic from the world’s waterways has led to a surge of research in computer vision and AUV solutions. A team of researchers at the University of Minnesota robotics lab recently experimented with AUV deployments for identifying deep ocean marine plastic debris \cite{fulton2018robotic}. Another growing trend has been to utilize deep learning and computer vision to identify floating marine plastic on the river automatically and ocean surfaces \cite{PANWAR2020100026}.
Additionally, AUV’s have been used as a means for environmental surveillance \cite{5509604}, mapping \cite{5603860}, and localization of marine plastic debris [20]. Underwater vision technology has been pushed forward thanks to work done by Ge et al. \cite{Ge_Shi_Mei_Dai_Li_2016} with LIDAR technology to localize and map marine-plastic debris on coastal beaches. Further research into implementing LIDAR in conjunction with forward-facing SONAR image models trained by deep convolutional neural networks was conducted by Howell et al. \cite{Kurz_Buckley_Howell_Schneider_2009}, and Valdenegro-Toro et al.
\cite{valdenegrotoro2019deep} which resulted in a model capable of detecting underwater debris with 80\% accuracy. Unfortunately, these methods incur high expenses due to retrofitting sonar and an in-house water tank for evaluation.
The University of Minnesota robotics lab \cite{fulton2018robotic} annotated and published a dataset of images collected by the Japan Agency for Marine-Earth Science and Technology (JAMSTEC) \cite{JAMSTEC}. JAMSTEC released the J-EDI (JAMSTEC E-Library of Deep-Sea Images), which contains marine plastic debris dating back to 1982 and provides data in the form of images and videos. The work presented in this research paper has benefited from the University of Minnesota team, which released close to 3000 annotated images from the JAMSTEC J-EDI dataset. These datasets were used to train our convolutional neural networks (CNNs) to identify features of plastic debris.
Photography, especially video-cameras, have found common application as environmental monitoring systems \cite{mock1995underwater} \cite{premkumardeepak2017intelligent}. Underwater cameras provide a globally accessible and low-cost quantification aid. Combining object detection models with underwater cameras equipped on automobiles such as AUVs makes it possible to observe and monitor sub-surface plastics in known hotspots worldwide \cite{fulton2018robotic}. By mounting video cameras to AUV’s, buoys, and other submersibles, institutions could feasibly quantify macro-plastics, which constitute 90\% of the total plastic mass in the oceans.

\

\begin{figure}[ht]
    \centering
\subfigure[Ocean]{\label{fig:ocean_plastic}\includegraphics[width=0.24\textwidth]{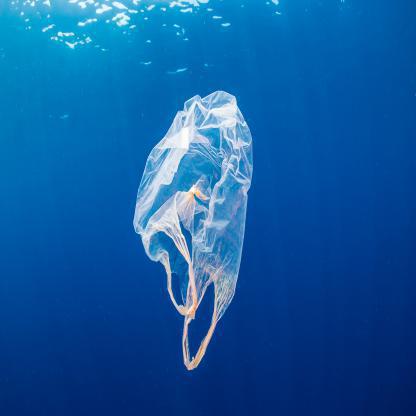}}
\subfigure[Lake]{\label{fig:lake_plastic}\includegraphics[width=0.24\textwidth]{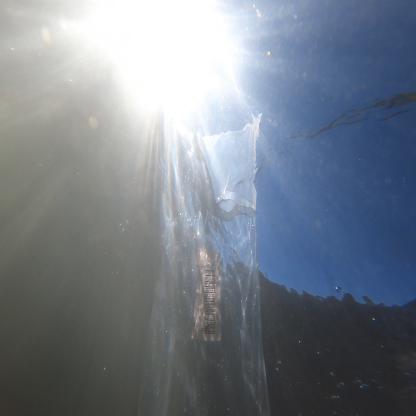}}
    \caption{Example images of marine plastic debris from the DeepTrash dataset in different marine environments}
\label{fig_plastic}
\end{figure}

\begin{figure*}[!t]
\centering
\includegraphics[width=0.95\textwidth]{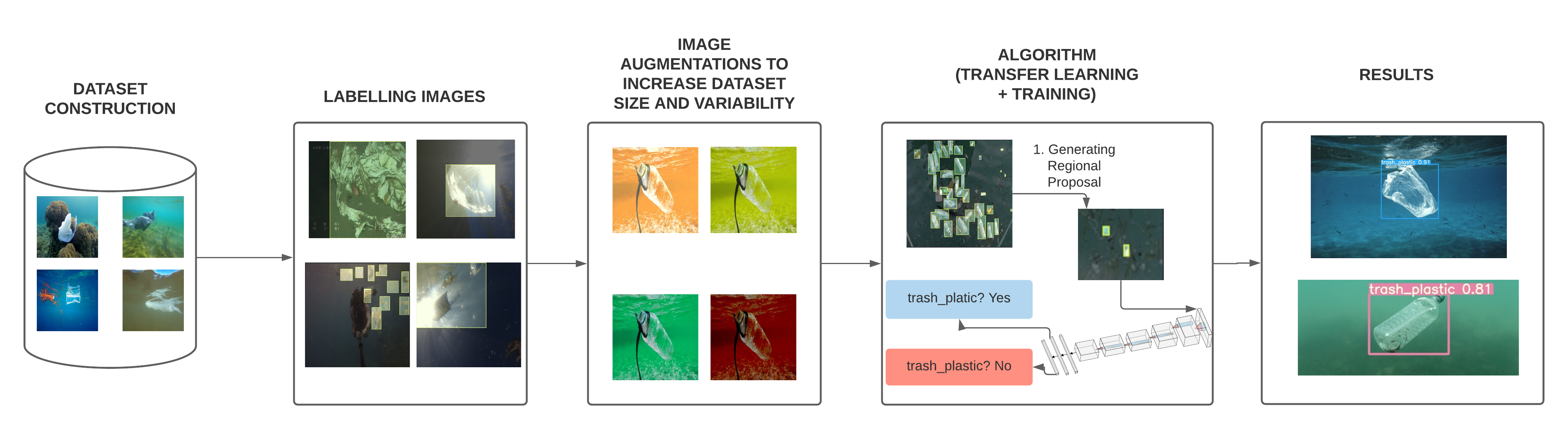}
\caption{Methadology for Marine Plastic Detection}
\label{fig_met}
\end{figure*}

\section{Network Architecture}

Four state-of-the-art object detection models were selected for this work. Each architecture has different benefits and drawbacks, with the main trade-off being speed for accuracy.


\begin{itemize}

\item \textit{Faster RCNN Inception v2}
Faster RCNN \cite{10.5555/2969239.2969250} is an improvement on R-CNN \cite{DBLP:journals/corr/GirshickDDM13} that
introduces a Region Proposal Network to make the network trainable end to end. The network uses the convolutional feature maps to produce
region proposals, which are fed to the fully connected (in our case softmax layers) for detection. Faster R-CNN uses VGG-16 \cite{DBLP:journals/corr/SimonyanZ14a} for feature extraction while we use the newer we use the Inception v2 \cite{tensorflowmodelgarden2020} as feature extractor instead because of it's known capabilities to enhance object detection.

\item \textit{Single Shot Multibox Detector MobileNet v2}
Single Shot MultiBox Detector (SSD) \cite{DBLP:journals/corr/abs-1805-09501} is another well-known
detection model that performs object localization and
classification in a single forward pass of the network. This architecture introduces additional Convolutional layers with the base network to improve performance. We use a MobileNetv2 implementation \cite{NAGRATH2021102692} for faster performance speeds. 

\item \textit{YOLOv5-S} \cite{Jocher_Stoken_Borovec_NanoCode012_ChristopherSTAN_Changyu_Laughing_Tkianai_Hogan_Lorenzomammana_et_al._2020}
YOLOv5-S Unlike the official release of YOLOv4, YOLOv5 currently exists in active development. There- fore, all YOLOv5 related code, and models may be subject to modification or deletion without notice. YOLOv5-S has 7.5 million parameters, 140 layers and operates at a lightweight 7MB (14MB for weights pre-trained on COCO). This architecture uses the Cross Stage Partial Network (CSP) \cite{wang2019cspnet} as the processing backbone and was trained on MSCOCO to extract rich/informative features from an input image. YOLOv5 also uses a PANet \cite{liu2018path} for the model-neck to generate feature pyramids and the computational friendly LeakyReLU and Sigmoid activation function. The model uses SGD as a default learning rate, but these tests were performed with the ADAM adaptive learning rate enabled \cite{kingma2017adam}.

\item \textit{YOLOv4-Tiny} \cite{bochkovskiy2020yolov4}

Inference speeds on YOLOv4-Tiny can reach upwards of 400 frames/second when using a 1080Ti GPU with accuracy, precision, and recall that meet the demands of a production-ready robotics platform. YOLOv4- Tiny uses a CSPDarknet53-Tiny neural network as opposed to the regular SPDarknet53 network. To simplify the computation process, the YOLOv4-Tiny model uses the LeakyReLU as an activation function.

\end{itemize}

\section{Methodology}

\subsection{Dataset Construction}

The dataset was curated by collecting videos of marine plastic from the field in California (South Lake Tahoe, Bodega Bay, San Francisco Bay). The videos vary significantly in quality, depth, and visibility to better represent the harshness of marine environments.

After recording, manual identification of marine plastic captured in the still images was performed, emphasizing choosing images containing complex object detection scenarios such as illumination, noise and occlusion. Each image would then get annotated to prepare them for object detection using the deep learning models. This curation approach ensured that the dataset of images would closely conform to real-world conditions.
To further increase the representation of marine plastics in different locations, images were also sourced from datasets created by the Japan Agency for Marine-Earth Science and Technology (JAMSTEC) \cite{JAMSTEC}. 
Annotations were performed using the free tool supervisely \cite{drozdov} to create the final dataset, which contains ~3200 images. Ocean environments provide a wide variety of visual challenges, so all plastic instances get consolidated into a single classification labeled “trash\_plastic”. We called our final dataset, DeepTrash \cite{tata_gautam_2021_5562940} and have open sourced it to further help the research in this field.

\subsection{Enhancements of Custom Dataset}

The following procedures were implemented for the deep learning models to detect marine plastic:

\begin{enumerate}[label={\alph*})]
\item \textit{Dataset Formatting} 
The input data constituted of images and annotation labels for bounding boxes were converted into either a TFRecords (FasterRCNN and SSD), PyTorch (YOLOv5-S) or a Darknet format (YOLOv4) to process each respective model. The bounding boxes delimited each image’s regions of interest based on 2D coordinates located in the respective annotation file.

\item \textit{Image Pre-processing} To ensure that learning occurs on the same image properties, auto orient was applied to strip images of their exchangeable Image file format (EXIF) data \cite{9108753} so that the models interpret images regardless of image format. Finally, the input images get resized and bounding boxes adjusted to 416x416 pixels.

\item \textit{Data Augmentation} To mitigate the effects of the model generalizing towards undesired features and to replicate underwater conditions such as variable illumination, occlusion, and color–the dataset was further enhanced by randomly changing the brightness and saturation of the images via PyTorch’s built-in Transforms augmentation. These modified images were then added back into the dataset, effectively tripling the size of our dataset.

\subsection{Object Detection}
We used four state-of-the-art neural network architectures FasterRCNN with Inception v2, Single Shot Multibox Detector with MobileNet v2, YOLOv5-S and YOLOv4, downloaded from their respective repositories . The following software versions were used: Tensorflow1.5, PyTorch v1.8.1, Darknet, OpenCV version 3.2.0, and CUDA 11.2.

\subsubsection{Fine Tuning Hyperparameters} 
This object detection model uses ADAM \cite{kingma2017adam} as the adaptive learning rate, which utilizes a decaying learning rate for a set number of epochs. The final layer of the network uses Softmax and reflects the usage of a single class.

\subsubsection{GPU Hardware}
We tested two state of the art GPUs:NVIDIA P100 and NVIDIA Tesla V100®GPU (version 460.32.03) were chosen due to their proven parallel computing capability.

\subsubsection{Training}
After every 1000 epochs (iterations) of training, the model would be evaluated on the validation dataset to calculate precision, recall, and mean average precision (mAP). This means stopping training to check for the following:

\begin{itemize}
\item When accuracy stops increasing, the model no longer needs additional training to prevent overfitting.
\item Depending on performance, hyperparameters should receive adjustments to optimize for evaluation metrics.
\end{itemize}

\subsubsection{Evaluation Metrics}
After the model has finished training, use the testing and validation datasets containing images mutually exclusive from the training dataset as an input to evaluate the network’s performance.
The model draws a bounding box around successfully detected objects with a confidence score of .50 or higher. The number of true positive bounding boxes drawn around marine plastic debris and true negatives provides the basis of evaluation. The following performance metrics were utilized to produce results:

\begin{itemize}
\item \textit {\textbf{True positive and True negative values}}: True positive values represent an outcome in which the models correctly predict a positive class, and conversely, a true negative represents when the model correctly predicts the negative class.

\item \textit{\textbf{Precision and Recall}} -- represents if the model suc- cessfully detected plastic in an image.
\[Recall = \frac{TP}{TP+FN}\]
\[Precision = \frac{TP}{TP+FP}\]
\item \textit{\textbf{Mean Average Precision}} -- Evaluates how often the network can recognize plastic in a group of images. After collecting the values for true and false positives, generate a precision-recall curve using the Intersection over Union (IoU) formula:
\[ \textit{IOU}=\frac{BBox_{predicted} \cap BBox_{groundTruth}}{BBox_{predicted} \cup BBox_{groundTruth}} \]
Where \(BBox_{predicted} \)
and \(BBox_{groundTruth}\)
Where BBoxpredicted and BBoxgroundT ruth are the areas under the curve for predicted and ground truth bounding boxes, respectively. A high threshold for confidence and IoU must be set to ensure accuracy, with a correct detection represented by the threshold being exceeded.

The mAP can then be obtained by integrating the precision-recall curve \cite{10.1007/978-3-642-40994-3_29}:

 \cite{10.1007/978-3-642-40994-3_29}:
\[  mAP = \int_{0}^{1} p(x) dx \]

\item \textit{\textbf{F1-Score}} -- Evaluates the balance between precision and recall values.

\item \textit{\textbf{GPU Speed (ms/IMG})} -- Represents how fast the network can infer marine plastic debris contained within an input image.

\end{itemize}

\subsubsection{Visualizing results}
For each processed image, the network populates arrays containing the following data: Scores -- Confidence scores for the predicted boxes and Number of detections -- The total number of detections made per image.

The following equation converts the normalized coordinates into image coordinates for rendering bounding boxes on top of images:

\begin{equation}
imgCoord_k = BoxScore_{i}^{j}\cdot Width
\end{equation}
where $k\in$~(left,right,top,bottom), $i$ is an index of boxes, $j \in(0,1,2,3)$, and $Width$ is a width of the image. These image coordinates were used to visualize the results of predicted bounding boxes in Figure \ref{fig_results}.

\end{enumerate}

\begin{figure}[ht]
    \centering
\subfigure[Detection near water surface]{\label{fig:near_water}\includegraphics[width=0.24\textwidth]{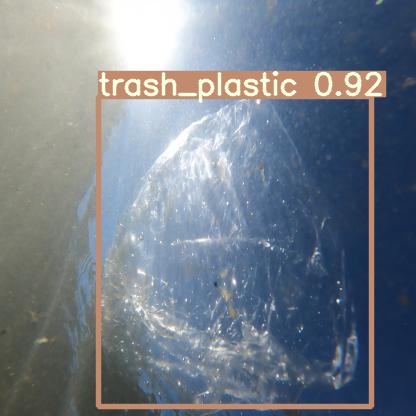}}
\subfigure[Detection for partially buried debris.]{\label{fig:partial_occlusion}\includegraphics[width=0.24\textwidth]{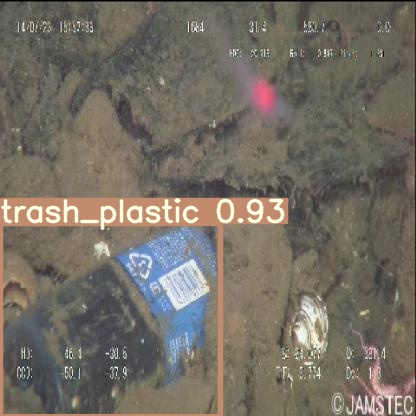}}
    \caption{Results generated by the model with bounding boxes and confidence scores rendered over marine plastic debris.}
\label{fig_initial_results}
\end{figure}

\begin{table}[t]
  \centering
  \vspace{3mm}
  \begin{tabular}{l|c|>{\centering\arraybackslash}m{0.7cm}|*{3}{c}}
    Network& mAP&F1-Score&Precision \\
    \hline
    YOLOv5s&85.0&0.89&\textbf{0.93}\\ 
    Tiny-YOLOv4&84.0&0.80&\textbf{0.96}\\ 
    Faster R-CNN&79.0&0.76&\textbf{0.84}\\
    SSD&76.0&0.71&\textbf{0.83}\\
  \end{tabular}
  \caption{Detection metrics in mAP, F1, and Precision.}
  \label{tab:detection_one}
\end{table}

\begin{table}
\centering
\begin{tabular}{  l | c | c | c }
Network & P100 & V100  \\
\hline
YOLOv5s &  2.8 & 1.4  \\
Tiny-YOLOv4 & 1.9  & \textbf{1.2}  \\
Faster R-CNN & 2.4 & 1.5  \\
SSD & 2.5 & 2.1  \\
\end{tabular}
\label{tab:fps}
\caption{Performance metrics for Inference (ms/img).}
\vspace{1mm}
\end{table}



\section{Results}

All results expressed in Table I were produced from the validation dataset presented in the methodology section. Since the images used in the training dataset were not isolated laboratory creations but instead real-world images directly from the field, the general object detection has a more accurate representation of marine plastic debris. This approach comes with a set of trade-offs:

\begin{itemize}
    \item The model performs stronger in real-world deployments, and therefore, the evaluation results in Table II do not significantly differ from near-real-time measurements taken from the field.
    \item Peak performance of the object detection model in a perfectly controlled environment could not be measured, and the highest possible benchmark of a single detection remains unknown.
    \item These trade-offs indicate the results of this paper better approximate long-term performance across a wider variety of marine environments–leading to a more substantial evaluation of the object detection model’s performance in the field.
\end{itemize}

\subsection{Quantitative Results}
The results captured in Table I demonstrate that near-real-time object detection of marine plastic debris in the epipelagic layer of the ocean is both feasible and close to real-world execution. The tested models demonstrate high average precision, mAP, and F1 scores relative to their inference speed. Repeated testing of the model produced a results variance of 2
Usually, evaluation results between models showcase a clear relationship between models, such as trading off significant inference speed for increased accuracy. However, the results presented in this paper showcase that both YOLOv4-Tiny and YOLOv5- S produce high debris localization metrics when it comes to identifying epipelagic plastic in near-real-time.
YOLOv5-S provides a significantly higher F1 score in
exchange for a slight dip in inference performance.
Reducing the number of classes to 1, i.e.,” trash plastic,” ensures an even distribution of class examples within the training dataset. The singular nature of this object detection model may reduce the total number of use cases the model can be utilized for–but guarantees strong performance on use cases within the domain of the model. A single classification also builds upon the pre-trained weights’ performance during transfer learning, as it meant less skewing towards unrelated classifications.

\subsection{Evaluation Results}

\subsubsection{Object Detection}
The mAP values obtained from the object detection models on the validation dataset have been expressed in Table I. Both models demonstrate high accuracy in plastic localization. It also reveals that the YOLOv5-S model has a higher mAP than the YOLOv4-Tiny, Faster RCNN and SSD models.

\subsubsection{Inference Speed}
These speeds were dictated by the GPU (NVIDIA P100 and V100 using a batch size of 32) and included image preprocessing. The YOLOv4-S model provided the highest inference speed-to-maP performance ratio for the provided dataset.

\subsection{Qualitative Results}
This study focused on determining the feasibility of detecting marine plastic debris for near-real-time monitoring/quantification purposes. To that end, the results in Table I demonstrate that general object detection models can fill this much-needed role. Since a relatively high level of performance can be maintained with such fast inference speeds–we believe that models such as the one presented in this paper can be applied to AUVs and other tools for real-world solutions. Equally important is that these solutions now have a near-future timeline of implementation and have been proven to be low-cost.


\section{Discussion}
\label{sec:disc}
In this study, we built a computer vision model that detects marine plastic debris with high precision, visualizes the detections with bounding boxes, and operates near real-time speeds. These conditions match the requirements for robotic platforms such as AUVs or buoys.
As one of the first object detection models specialized for the epipelagic layer, direct comparison results can not be readily performed—however, relative performance comparisons.

DeepPlastic and object detection models geared towards plastic detection in deep-sea and river plastic environments reveal DeepPlastics’ state-of-the-art performances.
The article mentioned above in Earth and Space Science \cite{https://doi.org/10.1029/2019EA000960} describes a two-stage reference model, utilizing cameras positioned above water, capable of detecting plastic floating on rivers. It utilized 1272 images in its training set and the Faster R-CNN architecture for its second stage. Across multiple experiments, this model’s highest precision rate was  68\% when employing image flipping and the ADAM adaptive learning rate. DeepPlastic was trained on a dataset using image flipping, in addition to other data augmentation techniques, and also uses the ADAM learning rate--but achieves a precision rate of 93\% when detecting marine plastic debris submerged in the ocean via underwater cameras.
The University of Minnesota’s (UoM) computer vision model \cite{fulton2018robotic} specialized for marine plastic detection in deep-sea environments utilized 5720 images in its training set and three classes. The DeepTrash training dataset shares many of the same images as both include samples from JAMSTEC \cite{JAMSTEC}. The UoM model achieved an mAP of 82.3\% for its plastic images class using the YOLOv2 architecture and a high of 83.3\% when using Faster R-CNN. DeepPlastic achieves an mAP of 93\% when using the YOLOv4 architecture and input images of marine plastic debris on the same dataset in similar conditions. 
AquaVision [17] was trained on three datasets totaling ~4400 images, including images of both land-based and marine-based debris and four classes. AquaVision’s highest performance for the plastic class was an average precision of 81.5\% when using the one-stage RetinaNet method. DeepPlastic performs at an mAP of 85\% when using YOLOv5.
The specific training datasets used by the three models described above are either not public or utilize datasets outside of the domain of DeepPlastic (i.e., the dataset images are not underwater). Therefore comparing performances via dataset is not an option for this study.

\subsection{Points of Improvement}
This model can efficiently monitor and quantify marine plastic. Improvements can be made in the following areas:

\subsubsection{Data Augmentation Improvements}
While grayscale, saturation, and vertical/horizontal flipping have been proven data augmentation techniques–emerging techniques such as AutoAugment [35] could be explored to improve the model’s variability in the future once ready for adaptation. Other methods such as shear and the cutout regularization technique would be great to utilize after integration technologies improve.

\subsubsection{Dataset Improvements}
The data set used in this study is unique and one of the first of its kind. For the data set, we see three main improvements that can be made to enhance the deep learning model:

\begin{itemize}
    \item Adding more images from different locations
    \item Using more images from variable water conditions
    \item Finally, acquiring a more extensive set of underwater plastic images
\end{itemize}

As more plastic images from different locations and oceanic conditions become available, they will increase marine plastic debris representation–providing a more comprehensive dataset for model training. We believe this will improve the mAP and overall robustness of the object detection model.

\subsubsection{Camera Improvements}
Readily available off-the-shelf cameras have come a long way but still suffer from certain limitations.
The first and most substantial limitation revolves around most underwater cameras that will only work during the daytime. If we want to continue the monitoring process during the nighttime, better night-vision underwater sensors need to be developed.
The second limitation stems from the common H.265 video compression techniques \cite{Lu_2019_CVPR} underwater cameras utilize to induce encoding artifacts. This impedes real-time detection by deteriorating the image quality. Developments in end-to-end deep learning video compression techniques \cite{Lu_2019_CVPR} could lead to solutions for this limitation once ready for implementation.


\section{Code and Dataset Availability}

All code/dataset and instructions to build and utilize the DeepPlastic object detection model can be found \href{https://zenodo.org/record/5562940#.YWSe39nMI-S}{ online via Zenodo -- DeepPlastic}.
\indent


\section{Conclusion}
\label{sec:con}
This work’s objective was to develop a deep learning vision model capable of consistently identifying and quantifying marine plastic near real-time. To attain this objective, a pair of general object detection models were constructed using two state-of-the-art deep learning models built for inference speed to measure which performed best.
\\
\indent
This study concludes that a marine plastic debris detection system based on the YOLOv5-S model would be fast, accurate, and robust enough to enable real-time marine plastic debris detection.
This study shows that effective object detection models can be constructed using readily available, pre-enabled GPUs for reasonable costs.
\\
\indent
Furthermore, the dataset created for and utilized by this general detection model demonstrates that massive, highly curated datasets can be used in conjunction with samples relative to the domain of object detection and web scraping to produce promising results. 

\textit{This computer vision system enables multiple deployment methods to detect/monitor marine plastic and allows researchers to quantify marine plastic debris without physical removal.}

\section{Future Work}
Improvement of the dataset would have the highest impact on performance but collecting additional images would require human labor in fieldwork or preprocessing. A technology capable of producing synthetic images containing marine plastic debris in an ocean environment could provide an automated solution to dataset creation. This could be accomplished with a two-stage autoencoder [37]. Object detection models trained on identifying jellyfish (or other objects similar to marine plastic debris) paired with a our object detection model could lead to a decrease in false positives.
Inference speed could be improved through specialized GPU technology or tailoring models towards specific higher power GPUs than used in this study.
An end-to-end video compression technique explicitly developed for near real-time object detection could lead to a better ratio of true positives to true negatives and an improved range on object detection.
Tailoring this object detection model for vision-equipped AUVs could result in automated identification and plastic removal devices capable of scalable deployment across large bodies of water, as shown in figure 1. Further optimizations could build in support for stationary monitoring devices such as buoys as well. We hope that such a system will facilitate scalable adoption by researchers and civilians to detect and clean up marine plastic.

\section{Acknowledgements}
We gratefully acknowledge the help and support of Nikhil Deshmudre for his efforts and help with the deployment of this computer vision system. 

The authors would also like to thank Joseph Nelson, Co-Founder of Roboflow.com, for providing us with Roboflow Pro free of charge, making it easier to iterate on the deep learning models. Some of the images in this dataset were sourced from the TrashCan dataset, where the researchers hand-annotated and open-sourced over 5000 images from the JAMSTEC-JEDI dataset. 

The authors would like to thank the researchers from the University of Minnesota, Robotics LAB, and the Japan Agency for Marine-Earth Science and Technology for open sourcing this data to contribute to the advancement of science. Finally, we would like to thank Rae Rose Lowe for her support throughout this process.

\appendix

\printcredits

\bibliographystyle{cas-names}

\bibliography{cas-refs}


\end{document}